# Leveraging Advancement in Ambient Technology for Effective Healthcare

Amos Okomayin
Department of Computer Science,
Middlesex University,
London, United Kingdom

Tosin Ige
Department of Computer Science,
University of Texas at El Paso,
Texas, USA

## 1.0 Introduction

Ambient intelligence refers to technologically enhanced electronic environments which are both responsive and sensitive to the presence of people within their environment. Environments that are integrated with ambient intelligence tends to adapt to the needs of individuals within the environment in an unobtrusive manner in such a way as to enhance everyday life thereby making interaction with technology extremely seamless and well-integrated. This capability was made possible because it is a concept that combines several key technologies such as IoT (Internet of Things) technology, sensor technology, AI (Artificial Intelligence), and advanced human-computer interaction all embedded and integrated together with the environment despite the sophistically and potency of cyberattack operation (Ige et al., 2024).

Today, we have a mixture of young and older individuals, people with special needs, and people who can care for themselves. Over 1 billion people are estimated to be disabled; this figure corresponds to about 15% of the world's population, with 3.8% (approximately 190 million people) accounting for people aged 15 and up (Organization, 2011). The number of people with disabilities is upward due to the increase in chronic health conditions and many other things. These and other factors have made the need for proper care facilities urgent in today's society. Several care facilities are built to help people with disabilities live their everyday lives and not be left out of the community.

## 1.1 What exactly are care facilities?

Care is caring for somebody or something and providing what they need for their health or protection (Oxford University Press, 2022). Moreover, *facilities* are defined as buildings, services, equipment, etc., provided for a particular purpose. Hence, care facilities can be seen as equipment or services used to provide health protection to someone or people in need.

In today's care homes, healthcare workers, such as doctors, nurses, midwives, etc., perform repetitive tasks such as monitoring their patients to ensure they do what they are supposed to or assisting. We are currently experiencing an increasing need for care workers. Some people travel miles to meet up with clients or patients. The National Health Service (NHS) faces a general shortage of about 100,000 healthcare workers across the health sector (Fund, 2018). If current trends continue, the need will be around 250,000 by 2030. In a media release, 40% of medical doctors are close to retirement age in one-third of Europe and Central Asia countries (WHO, 2022). The workforce in this sector is declining, hence the need for a system that can silently assist in providing first-level care.

Imagine a system that can monitor and keep track of your health status continuously, diagnose possible health conditions, and offer advice on your habits to persuade you to do or stop some things you are currently doing that are detrimental to your health.

This system could be knitted into a regular clothing fibre or a wearable with a tiny sensor that uses ambient intelligence to communicate with various devices in your apartment, collecting data and giving instructions to all of these other devices in real-time.

The term "ambient intelligence" refers to the future vision of intelligent computing where the environment supports the people inhabiting it (E. Aarts et al., 2009). In this technology, the human effort used when inputting and outputting information will no longer exist; it depends on interconnected sensors communicating with one another to make critical decisions based on the data gathered.

## 2.0 System Engineering Features

2.1 Persona

**2.2.1 Bola's General information**

| **Bola, Canada** | | |
|---|---|---|
| **General information** | **Condition** | **Technology** |
| **Age:** 65<br>**Gender:** Male<br>**Family/support:** no family member<br>**Living Conditions:** Bola lives in a care home with three other men in their 60s80s with schizophrenia and autism. **Hobbies/Interests:** Bola enjoys listening to music and watching football matches<br>**Attitude/Feelings:** Bola is a confident and outspoken person. | **Health Condition:** High Blood pressure and Kidney failure<br>**Sensory:** Poor visual – needs to wear glasses. **Body:** Average weight.<br><br>**Skills:** Can play the keyboard.<br><br>**Limitations:** Bola is not very committed to taking his drugs. He needs to be prompted. | **The technology used:** Bola is not very conversant with technological gadgets. He has an iPad, which he can operate quite well. He needs constant assistance.<br><br>**Attitude towards technology:** Bola is not able to navigate his way very well across the internet |

## 2.2.2 Scenario Bola

**User type:** A person with Kidney failure and high blood pressure living in a care home need technology to closely monitor his vitals and ensure medication is taken at the right time.
**Actors:** James (50) and Support workers.
**Help needed with:** monitoring vitals, information processing and time management
Bola lives in a shared apartment in a care home with three other male friends who are autistic in North West London. He moved into the care home at age 60 after almost dying from cardiac arrest. He loves to be in the midst of people, but he has no family members, so he moved into a care home.

In his shared apartment, Bola has graphical schedules, and his care provider checks on him to ensure he has done all he needs to do daily. Bola needs something to remind him to take his drugs on time constantly, and just as his support worker does to him, he gets lost in time easily. This is due to lots of thinking which his support worker also tries to reassure him of **Bola's main problem is** the high blood pressure that comes up frequently which
Bola needs help in knowing what to eat and what not to eat to prevent an attack
He needs to be reminded of football matches and social activities

### 2.2.3 Identifying stakeholders for Bola

| Stakeholder | Description |
|---|---|
| Primary user | People with High Blood pressure and kidney |
| Secondary User | Carers of a patient with a medical condition in old age |
| Tertiary User | No tertiary User |
| Calls Provider | Companies that connect mobile devices with calls and SMS |
| Internet provider | Internet service provider for mobile devices |
| Device Manufacturer | Organisations that make the product |

### 2.2.4 Identifying Activities for Bola

| Stakeholder | Goals | Sub-Goals | Requirements |
|---|---|---|---|
| Primary User (PU) | Increase their independence from SUs on reminders to take medication and monitoring changes in their body system | Guide the PUs through reminders of what they need to do next and advise them on what to eat and what not to | Receive data(High Blood pressure and kidney status) from PU's body system and make decisions based on the received data |
| | | | PU can access his data and the analysed result of the data. |
| Secondary User (SU) | Reduce attention PU requires | Show that PU is safe | SU Can get all the vital body readings of PU remotely. |
| | | | |
| | | Show SU that PU is doing what he should do at the right time. | SU can know the current medical condition of PU and if he takes his medication constantly. |

### 2.2.5 Situational services and interaction types for Bola

| Activity | Situation of Interest | Situational Need | Situational Service |
|---|---|---|---|

| Take medication | It is time for the user to take medication | A record that the user has taken medication | Sensor Updates the state of the current activity |
|---|---|---|---|
| Vital readings | The carer needs users' data for analysis. | Take the readings of the user's body, kidney status and blood pressure. | Sensor updates the state of current activity |

### 2.3.1 James' General information

| **James, US** | | |
|---|---|---|
| **General information** | **Condition** | **Technology** |
| **Age:** 70<br>**Gender:** Male<br>**Family/support:** Has a son in Dubai<br>**Living Conditions:** Lives alone in a retirement home.<br><br>**Hobbies/Interests:** James enjoys watching football matches<br>**Attitude/Feelings:** James is a confident and introverted person. | **Health Condition:** Asthmatic<br>**Sensory:** Poor visual – needs to wear glasses. **Body:** Average weight. He has difficulties making healthy eating choices.<br><br>**Skills:** Can play the keyboard.<br><br>**Limitations:** James is not able to see very well without his glasses. He cannot tell if the environment hits high pollen, which could expose him to an attack. | **The technology used:** James is not very conversant with technological gadgets.<br>He has a smart inhaler.<br><br>**Attitude towards technology:** James love technology and believes it is here to improve our lives. |

### 2.3.2 Scenario James

**User type:** A person with Asthma at retirement age
**Actors:** James (70), hospital and Support workers.
**Help needed with:** ensuring the room is ventilated, needs help during asthmatic attacks and monitoring the respiratory system.
James lives alone in a retirement settlement. He moved into the home at age 60. He loves to do things alone.

In his apartment, James has graphical schedules; his care provider assists him with household chores and assistance during asthmatic attacks.
James is growing old and unable to move around as fast as he used to, so he sometimes needs help with some little things in the house.
**James' main problem is** the asthmatic attack that comes up frequently; He needs help knowing the state of his environment to prevent asthmatic attacks

### 2.3.3 Identifying stakeholders for Bola

| **Stakeholder** | **Description** |
|---|---|

| Primary user | People with Asthma. |
|---|---|
| Secondary User | Home carers, service support workers, and doctors. |
| Tertiary User | Son |
| Calls Provider | Companies that connect mobile devices with calls and SMS. |
| Internet provider | Internet service provider for mobile devices. |
| Device Manufacturer | Organisations that make the product. |

### 2.3.4 Identifying Activities for Bola

| **Stakeholder** | **Goals** | **Sub-Goals** | **Requirements** |
|---|---|---|---|
| Primary User (PU) | Increase their independence from SUs by checking tendencies of | Helps the PU ventilate the environment at home and advises him to leave high pollen or polluted environment. | Read the amount of air entering the PU's system and its content and make decisions based on the received data. |
| | asthmatic attacks and ventilating the room. | Advises the user what to do during an attack and assist as best as possible | PU can access his data. |
| Secondary User (SU) | Reduce attention PU requires. | Show that PU is safe. | SU Can get all the respiratory readings of PU remotely. |
| | | Show SU that PU is doing what he should do at the right time. | SU can know the risk of an attack based on data received from the body. |

### 2.3.5 Situational services and interaction types for Bola

| **Activity** | **Situation of Interest** | **Situational Need** | **Situational Service** |
|---|---|---|---|
| Leave the environment for better ventilation. | It is time for him to leave his current environment. | A record that the user has left the environment. | Sensor updates the PU's response. |

| Respiratory readings. | The system hospital/ carer needs to keep track of the PU's respiratory reading. | Take the respiratory reading of the PU's body. | Sensor updates the current activity. |
|---|---|---|---|

### 2.4.1 James' General information

| **Joy, UK** | | |
|---|---|---|
| **General information** | **Condition** | **Technology** |
| **Age:** 2 months old<br>**Gender:** Female<br>**Family/support:** Mother and Father<br>**Living Conditions:** Lives with parents.<br><br>**Hobbies/Interests:** sleeping and eating<br>**Attitude/Feelings:** nothing | **Health Condition:** Down's syndrome<br>**Sensory:** Good visuals.<br>**Body:** Average weight.<br><br>**Skills:** No skills.<br><br>**Limitations:** Joy is not able to communicate | **The technology used:** No technology.<br><br>**Attitude towards technology:** No attitude. |

### 2.4.2 Scenario Joy

**User type:** A baby that needs monitoring
**Actors:** Joy (2 months old), her Mother and a doctor.
**Help needed with:** ensuring that Joy is safe at night.
Joy is a baby who needs monitoring.
**James' main problem is** that the mother sleeps a lot and feels the baby needs to be monitored and catered for at all times.

### 2.4.3 Identifying stakeholders for Bola

| **Stakeholder** | **Description** |
|---|---|
| Primary user | Babies |
| Secondary User | Nursing mother |
| Tertiary User | Doctor |
| Calls Provider | Companies that connect mobile devices with calls and SMS. |
| Internet provider | Internet service provider for mobile devices. |

| Device Manufacturer | Organisations that make the product. |
|---|---|

**2.4.4 Identifying Activities for Bola**

| **Stakeholder** | **Goals** | **Sub-Goals** | **Requirements** |
|---|---|---|---|
| Primary User (PU) | Ensure baby is safe at night<br>Prompt the SU to wake up | Check PU’s vitals. | Prompt the mother when the PU is crying |
| | | | SU can access PU’s data. |
| Secondary User (SU) | Allows SU to do other things. | Show that PU is safe. | SU can get her office work done remotely while taking care of the baby. |
| | | | SU can know when to take PU for a check-up. |
| | | Show SU that PU is doing what he should do at the right time. | |

**2.4.5 Situational services and interaction types for Bola**

| **Activity** | **Situation of Interest** | **Situational Need** | **Situational Service** |
|---|---|---|---|
| PU is crying | Mother gets notified | SU should care for PU, maybe breastfeed. | Sensor records event. |
| PU’s vital is not Ok. | SU gets notified and can. | Take PU to the hospital for a check-up. | PU need to be diagnosed. |
| PU is due for a check-up | SU gets a reminder before the due date and on the due date. | SU takes PU to the hospital | Sensor records event. |

**3.0 Expected behaviour**

A sensor is embedded in the user’s body or put on a wearable; this device can read body vitals in real-time, as described in Figure 3.0 below.

**System architecture and diagram**

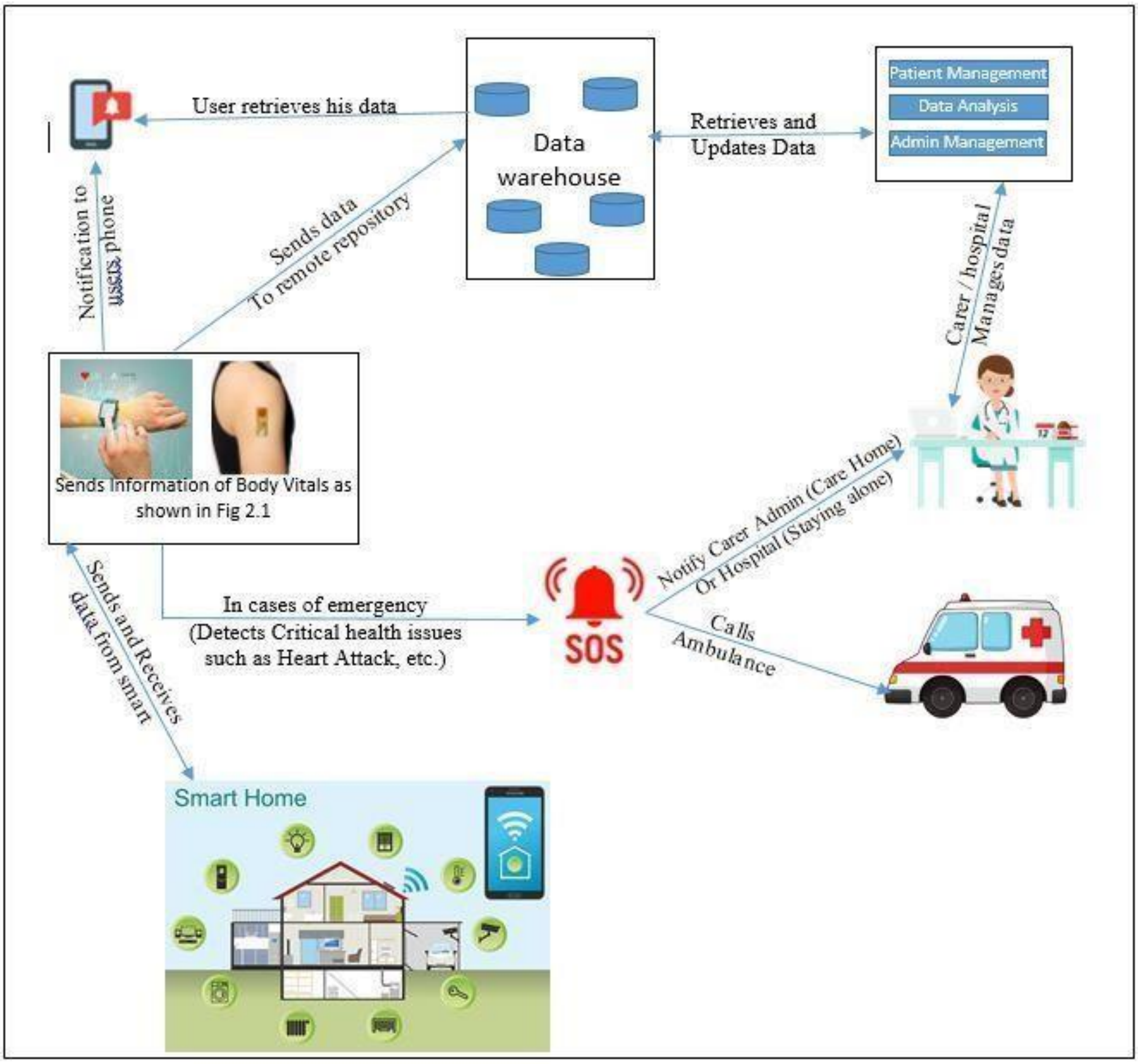

Figure 3.0. Showing the architecture of the System

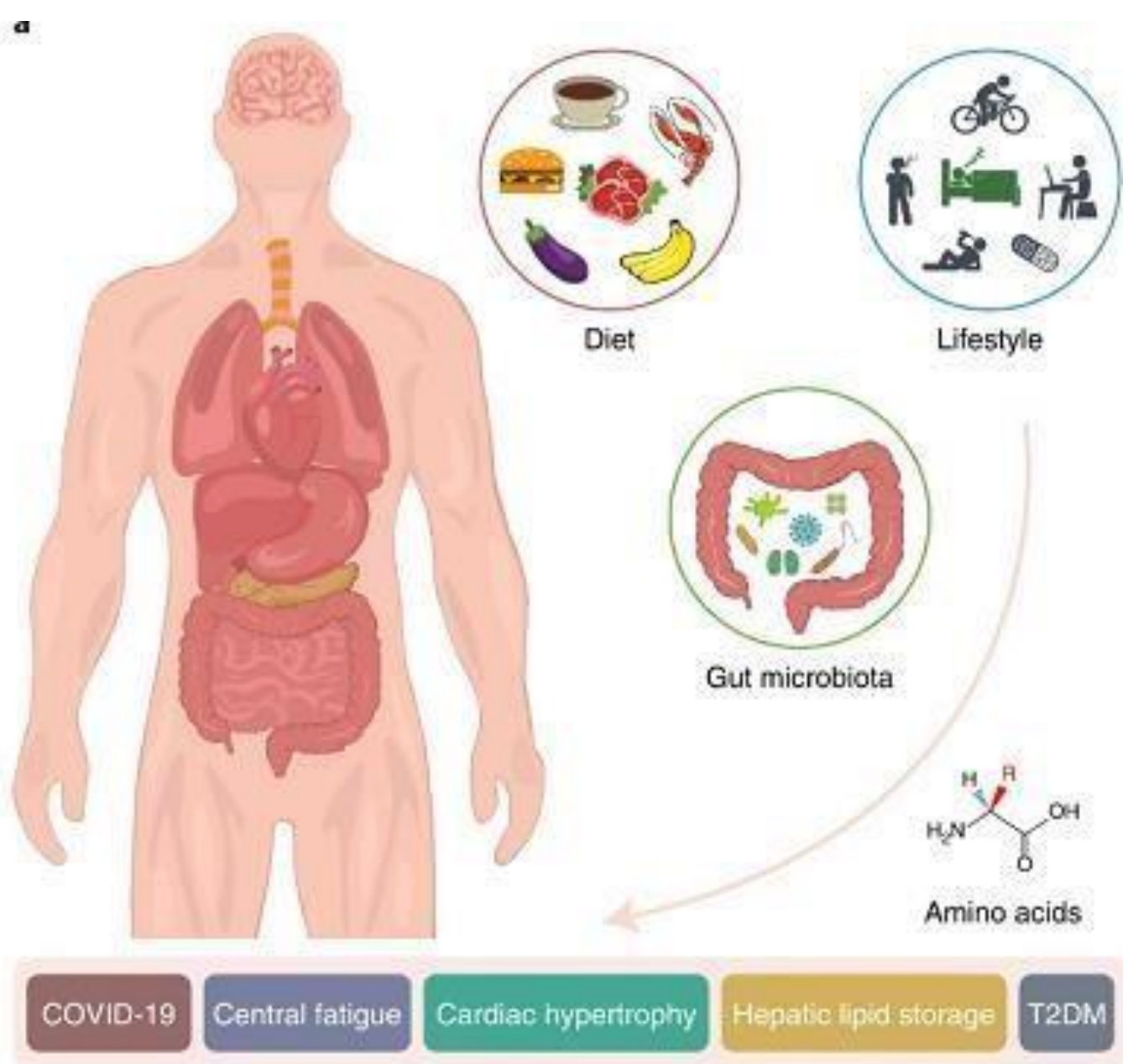

Figure 3.1. Showing the vitals the sensors are reading

The diagram above shows how the sensor communicates with the environment to ensure the user's safety.
The device is worn or embedded in the user's skin or knitted into a regular clothing fibre with a tiny sensor. This sensor reads the body vitals shown in figure 3.1 above, analyses them, makes decisions, and saves the details in a remote repository. It also communicates with various smart devices in the apartment, collects data from them, and gives them instructions to perform; all these are done in real-time.
This sensor has the following essential characteristics, Context awareness, Personalisation, Anticipatory, Adaptive, Ubiquity and Transparent.

**3.1.0 Describing Scenario for Bola and system response**

**3.1.1. Wake-Up Scenario and expected behaviour of the system**

- If it is time to take wake up system triggers an alarm for the user to wake up
- If the sensor senses no movement, the user's vitals are analysed for a liveliness check
- If the user's body readings/ vitals are nuanced, the system prompts the carer to wake the patient up
- If any of the user's vital is not OK, an SOS call is sent, notifying the nearest ambulance service and the carer for necessary medical care.

**3.1.2 Body Reading Scenario for Bola**

The sensor constantly takes readings of the kidney status and the blood pressure level of the user as all other necessary body readings.

- If the system notices any abnormality in Bola's body reading, such as his BP or kidney, the carer is prompted to attend to Bola
- If the situation gets worse, the system makes an SOS call to the nearest ambulance service

**3.2.0 Describing Scenario for James**

**3.2.1 Leave the current environment (Outside the house)**

- If the air in the environment hits,
- A high pollen rate or is polluted
- The sensor notifies the user to leave the environment and records the user's response

**3.2.2 Asthmatic Attack in the house**

- The sensor sends a signal to the windows to open up, - turns on the humidifier and
- Advises the user to use his inhaler.

**3.3.0 Describing Scenario for Joy**

**3.4.1 Joy is crying**

- If baby is cry
- prompt phone to wake the mother up,

**3.5.2 Joy's vital is abnormal**

- If Joy's vital is abnormal
- Prompt mother to contact the doctor.
- Ensure mother contact the registered doctor on the mobile app

**4.0 Societal Implications**

This list some critical societal implications and how the system addresses them.

1. Non-Maleficence / Beneficence: The users may be exposed to the infinitesimal amount of electromagnetic rays emitted from sensors; long-time exposure to them can lead to health hazards. An improvement in the system should be implemented to tackle this limitation.
2. User-Centred: The system was built to be customisable for each user's health-related issues, giving the user complete control of the system so the system cannot override the user's decision.
3. Multiple Users: The system must be built to accommodate different users with different roles and privileges to control what each user can see and do.
4. Privacy: Information shared will be consensual, the patient must agree to share data before it is shared, and a case of unintended data sharing is not possible according to the system design. Each user's data is discrete and is not transmitted or used outside the purpose for which It was collected.
5. Data Protection: All the sensor data collected are in line with countries' data protection policies. The system stores users' data securely in a distributed database so that there may be no loss of data and the system works effectively when a few data sources are down.
6. Security: The sensors communicate over TLS 3.0 protocols, the current standard of encryption, to ensure that a malicious user, which will lead to several risks, is not manipulating data.

7. Autonomy: These sensors are very much configurable, and users can override its actions.

8. Transparency: There is a precise term stating the nature of the data being stored, the people who can access it, the user being able to see his data and the side effect of the technology. This aims at providing transparency of data.

9. Equality, Dignity and Inclusiveness: The values and differences of the users are respected regardless of their sexual orientation, gender, disabilities, race, religious beliefs, political status etc. The cost of these sensors is quite affordable, considering the amount people put into care services.

## 5.0 Conclusion

The sensor connecting with smart devices in the home can tackle the shortage of caregivers and improve healthcare systems for people in need of care using Ambient Intelligent technology. It promises accuracy in real-time analysis of data, intelligently working in the background with minimal human input needed according to how it is configured.